\documentclass{article}

\usepackage{arxiv}

\usepackage[utf8]{inputenc} 
\usepackage[T1]{fontenc}    
\usepackage{hyperref}       
\usepackage{url}            
\usepackage{booktabs}       
\usepackage{amsfonts}       
\usepackage{nicefrac}       
\usepackage{microtype}      
\usepackage{lipsum}		
\usepackage{graphicx}
\usepackage{natbib}
\usepackage{doi}

\usepackage{eso-pic}%
\usepackage{fancyhdr}%

\usepackage{amsmath,amsfonts,bm}

\def\eqref#1{equation~\ref{#1}}

\def\1{\bm{1}}

\DeclareMathAlphabet{\mathsfit}{\encodingdefault}{\sfdefault}{m}{sl}
\SetMathAlphabet{\mathsfit}{bold}{\encodingdefault}{\sfdefault}{bx}{n}

\def\sB{{\mathbb{B}}}

\def\sH{{\mathbb{H}}}

\def\sM{{\mathbb{M}}}
\def\sN{{\mathbb{N}}}

\newcommand{\E}{\mathbb{E}}

\newcommand{\R}{\mathbb{R}}

\usepackage{natbib}%
\usepackage{comment}%
\usepackage[ruled,longend]{algorithm2e}%
\usepackage{hyperref}

\title{Spatially and Seamlessly Hierarchical Reinforcement Learning for State Space and Policy space in Autonomous Driving}

\date{} 					

\author{{Jaehyun Kim}
	\And
	{Jaeseung Jeong}
}

\renewcommand{\shorttitle}{\textit{arXiv} Template}

\hypersetup{
pdftitle={A template for the arxiv style},
pdfsubject={q-bio.NC, q-bio.QM},
pdfauthor={David S.~Hippocampus, Elias D.~Striatum},
pdfkeywords={First keyword, Second keyword, More},
}

\begin{document}
\maketitle

\begin{abstract}
Despite advances in hierarchical reinforcement learning, its applications to path planning in autonomous driving on highways are challenging. One reason is that conventional hierarchical reinforcement learning approaches are not amenable to autonomous driving due to its riskiness: the agent must move avoiding multiple obstacles such as other agents that are highly unpredictable, thus safe regions are small, scattered, and changeable over time. To overcome this challenge, we propose a spatially hierarchical reinforcement learning method for state space and policy space. The high-level policy selects not only behavioral sub-policy but also regions to pay mind to in state space and for outline in policy space. Subsequently, the low-level policy elaborates the short-term goal position of the agent within the outline of the region selected by the high-level command. The network structure and optimization suggested in our method are as concise as those of single-level methods. Experiments on the environment with various shapes of roads showed that our method finds the nearly optimal policies from early episodes, outperforming a baseline hierarchical reinforcement learning method, especially in narrow and complex roads. The resulting trajectories on the roads were similar to those of human strategies on the behavioral planning level.
\end{abstract}


\section{Introduction}
The essential idea of hierarchical reinforcement learning (HRL) is to find a proper hierarchy of abstractions for tasks on loosely coupled Markov decision process (MDP) \citep{parr1998reinforcement}. One predominant technique is temporal abstraction of policies, in which high-level command operates on Semi-MDP that has lower temporal resolution than MDP, and the low-level policy specifies primitive actions more frequently under the command \citep{sutton1999between}. Another technique is state abstraction, for which similar concepts are suggested in \citet{DBLP:conf/nips/DayanH92}, \citet{ DBLP:conf/nips/SinghJJ94}, and \citet{dietterich2000hierarchical}. Regarding the two types of abstraction, \citet{sutton1999between} proposed option framework, where high-level policy selects an option that consists of an initiation set, a sub-policy, and a termination condition. In respect of rewarding, two major directions exist: reward hiding \citep{DBLP:conf/nips/DayanH92} empowers the manager to independently reward sub-managers according to their compliance with the commands, while MAXQ value decomposition \citep{dietterich2000hierarchical} assigns credit for task rewards by temporally decomposing a state-action value in the parent task into expected total rewards during the execution of sub-policy and after the execution of sub-policy.

Advances in hierarchical approaches with deep neural networks, rooted in the aforementioned techniques, engage with multiple challenges. \citet{NIPS2016_f442d33f} presented hierarchical deep Q networks where every policy is learned as a separate DQN and intrinsic rewards encourage sufficient exploration in the subtask. For tasks on MDPs that are densely coupled in temporal dimension, high-level goals are modified by a network module \citep{DBLP:conf/nips/NachumGLL18}. For discrete sub-policies, option-critic \citep{DBLP:conf/aaai/BaconHP17} learns the options end-to-end by policy gradient, without having to manually define options. More diverse strategies became available through the use of a maximum entropy objective that enables a latent layer of high-level policy networks to directly control the sub-policy \citep{DBLP:conf/icml/HaarnojaHAL18}. HiPPO \citep{DBLP:conf/iclr/LiFCA20} adopted the control, formulated an unbiased latent dependent baseline, and derived a new hierarchical policy gradient that allows joint training of all levels with proximal policy optimization (PPO).

Recent approaches of HRL for autonomous driving (AD) \citep{DBLP:conf/iros/PaxtonRHK17, DBLP:conf/iros/ChenDPMMD19, DBLP:conf/itsc/RezaeeYNAEL19} kept up with some of these advances but did not fully reckon with following task-specific attributes. In AD, vehicle states can change unpredictably, rapidly, and frequently due to behaviors of ego agent and others. This does not satisfy the prerequisite of canonical HRL, a loosely coupled MDP. Additionally, the agent must drive through obstacles including other agents and avoid collision. Otherwise, it will be given a huge penalty and encounter the end of an episode without reaching the goal that accompanies delayed positive rewards. Therefore, the agent is demanded to keep an appropriate distance from obstacles concerning expected returns. In this respect, one way to efficiently learn navigation in AD is to sample the short-term goal only in collision-free regions which are relatively safe in near future and modify or re-sample them at subsequent time steps.

Even when given the knowledge about collision-free regions, the optimization is still tricky with gradient-based methods when the structure of regions is complex. In dense traffic, locally optimal intermediate goal position may exist for every inter-vehicle region (IVR) on a lane due to the contradiction of policy gradient directions between huge and hard-constrained objectives about collision avoidance and small and soft-constrained objectives about travel time reduction. On this ground, choosing one region among reachable regions in near future can be an effective task decomposition where the subtask is to find the local optimum in the given IVR. 

Thus, we propose a reinforcement learning method with a two-level hierarchy that accommodates those attributes of AD on highways. Our agent can estimate the value of multi-lateral strategies including short-term goals that can be local optima in IVRs and select the best one given the circumstance at every time step. Our contribution to HRL and AD is three-fold: 

We rethink the role of hierarchical policies, and propose a spatially hierarchical reinforcement learning (SHRL) method using deep neural networks. Our high-level policy selects a combination of behavioral sub-policy and its components, the IVRs to be used as a part of state for the two levels and as the outline of the sub-policy space. Our low-level policy generates an action in continuous space by elaborating within the outline, using a unique network structure and algorithm that we name neural elaboration within reinforced outline (NEWTRO).

We suggest a seamless HRL method for smooth state transition in short-term goal planning. The high-level policy selects one of either the current subspace or a candidate goal subspace from their subspace features, and the low-level policy can bridge the two subspaces. By sharing the value network for the two-level policies, our network structure and memory are designed to be as concise as those of a single-level actor critic, and learned without temporal abstraction.

Our methods showed drastic improvement of performance with fast optimization speed. In experiments, the agent using our method received higher rewards from early episodes than a baseline HRL, especially on narrow and complex roads. The traveling trajectory of the agent using our method followed general human tactics, while the agent using the existing HRL did not.

Two substructures of our methods and a conventional HRL method are compared on various types of roads and traffic volumes to examine how much effective to use collision-free regions that are the decomposition of space depending on obstacles.
\\
\section{Background and related work}
In this section, we introduce task definition in several branches of reinforcement learning (RL) and advantage actor-critic underlying our method to help understanding it. Then we compare our method in comparison with other HRL methods for autonomous driving on highways.

\subsection{Task definition in branches of reinforcement learning}
Canonical RL consists of the agent making action $a \in \mathcal{A}$ and the environment of the state $s \in \mathcal{S}$, interacting with each other. The environment follows MDP: it assumes the situation that given the current state $s_{t}$ at time $t$, the next state $s_{t+1}$ does not depend on the past states and actions. When the state is fully observable, the policy $\pi$ can be established from the $s$ to make $a$. In multi-agent RL, MDP is expanded to the actions of multi-agents ${a}$, where the next state is determined by $s$ and ${a}$. In certain RLs with feature set or hierarchical RL, $s$ can be abstracted to parts of $s$, or abstracted states can be defined in different space from $\mathcal{S}$. For goal-based RL methods of Universal Value Function Approximators \citep{DBLP:conf/icml/SchaulHGS15} or the policies given goal from the high level in hierarchy, the state-goal value $V(s,g)$ or the state-action-goal value $Q(s,a,g)$ can be estimated and learned. In single-agent RL or multi-agent RL with self-interested agents, a typical objective is the expected long-term discounted rewards $\rho(\pi) = \E \{ \sum_{t=1}^{\infty} \gamma^{t-1} r_{t} | s_{0}, \pi \}$. In RL with multi-objectives, the reward can be multi-dimensional, expressible as $r_{t} = \{r_{1,t}, r_{2,t}, \cdots, r_{N-1,t}, r_{N,t}\}$ for $N$ number of objectives. In this paper, we design a spatially hierarchical RL agent receiving a long-term goal and multi-objectives who operates on the environment with self-interested multi-agents.

\subsection{Policy gradient and advantage actor-critic}
Policy gradient methods aim to find optimal parametrized policies by performing gradient descent to optimize an objective.
The policy gradient theorem of \citet{DBLP:conf/nips/SuttonMSM99} derives the gradient with respect to the parameters of stochastic policy $\theta$ as

\begin{equation}
\frac{\partial \rho(\pi_{\theta})}{\partial \theta} = \underset{s}{\sum} d^{\pi_{\theta}}(s) \underset{a}{\sum} \frac{\partial \pi_{\theta}(s,a)}{\partial \theta} Q^{\pi}(s,a),
\end{equation}

where $d^{\pi_{\theta}}(s)$ is the stationary distribution following the parameterized policy $\pi_{\theta}$, and $Q^{\pi}(s,a)$ is the value of a state-action pair given a policy $\sum_{t=1}^{\infty} \E \{\gamma^{t-1} r_{t} - \rho(\pi) | s, a, \pi\}$.

For a designated start state $s_{0}$, only the long-term rewards are cared for $Q^{\pi}(s,a)$. In \citet{DBLP:phd/ndltd/Konda02}, $Q^{\pi}(s,a)$ is designed as a feature vector $\phi_{\theta}$, which is critic. Advantage actor-critic methods employ the advantage function $A^{\pi}(s,a) = Q^{\pi}(s,a)- V^{\pi}(s)$, where $V^{\pi}(s)$ is the state value used as the base line. The underlying reason is that the policy gradient suffers from its high variance because of the dependency on the state-action values during a trajectory. In our method we adopt n-step advantage estimate \citep{schulman2015high} because it fits our hierarchical actor-critic design aiming at efficient policy decomposition for domain-specific subspaces.

\subsection{Hierarchical policies for autonomous driving on highways}
Reinforcement learning approaches have been adopted for autonomous driving on highways to teach a vehicle agent several behaviors, such as cruise control \citep{DBLP:conf/ivs/ChenZLGW17, DBLP:journals/cim/ZhaoXZ17}, lane keeping \citep{DBLP:conf/icra/KendallHJMRALBS19}, lane changing \citep{DBLP:conf/ivs/WangCF18}, and traffic merging \citep{DBLP:conf/itsc/WangC17}. Recently, \citet{DBLP:conf/iros/PaxtonRHK17, DBLP:conf/itsc/RezaeeYNAEL19, DBLP:conf/iros/ChenDPMMD19} 
suggested hierarchical methods where the high-level policies commonly function as a type of behavior planner,
while the low-level policies take different roles: primitive control \citep{DBLP:conf/iros/PaxtonRHK17, DBLP:conf/iros/ChenDPMMD19} and motion planning \citep{DBLP:conf/itsc/RezaeeYNAEL19}.

The method of \citet{DBLP:conf/iros/PaxtonRHK17} generates motion plans by using tree search, where the high-level options and the low-level actions are selected in turn to extract the best option sequence. For the low-level policy, world states are abstracted to the set of continuous, logical, and agent-relative features. \citet{DBLP:conf/itsc/RezaeeYNAEL19} designed a cruising method on multi-lanes, where the high-level behavioral planner makes a decision on a discretized state-action space. They regarded path planning as a sequence of symbolically punctuated behaviors, and handed over authority of reactive planning to the low-level policy. They were concerned about specifying deadlines of a fixed expiration time of the high-level command in AD, so designed the low-level policy as a motion planner that prioritizes safety over the high-level decisions. \citet{DBLP:conf/iros/ChenDPMMD19} designed networks with spatial and temporal attention mechanism for input images of the front view. While a CNN encoder is shared for actor and critic, the spatial and temporal attention is applied only for hierarchical policies. On top of the encoder, salient regions are chosen by spatial attention networks, and the sequence of the regions runs through LSTM, whose outputs are applied with temporal attention.

Our HRL method does not take advantage of temporal abstraction, and is limited to short-term goal planning. Putting aside the concerns on the temporal aspect in \citet{DBLP:conf/itsc/RezaeeYNAEL19} and \citet{DBLP:conf/iros/ChenDPMMD19}, our method focuses on hierarchical policies with spatial decomposition because we consider that for driving on highways with dense traffic, spatially decomposing states can be much easier than temporally decomposing a sequence of multi-dimensional states that are entangled with hierarchical actions of multi-agents. For state abstraction, rather than using popular approaches of abstraction for sub-policy or spatial attention networks, we aimed at the high-level policy that can choose its own features resembling our eyes looking at several focus points and deciding to return to the salient one. While the hierarchical policies of previous research, the hierarchical policies have respective roles in different levels, such as abstracted behavior, motion planning, and control. our two-level policies work in the same policy space, which is the short-term goal generation.
\\
\section{Algorithm}
In the temporal respect, if a short-term goal is reachable without the episode ending midway, the expected returns of choosing a short-term goal position are the sum of expected returns during a behavioral motion before reaching the goal and expected returns at the future state when ego vehicle reaches the goal position. Although other agents are highly unpredictable, learning through multiple explorations enables estimating the expected returns of the goal selection given a state $s_{t}$, a long-term goal region $l_{t}$, and a behavioral motion to reach its short-term goal $g_{t}$ at time $t$. The details of a long-term goal region are in Section \ref{appendix:other_state_features} of Appendix, and we fix the region as $l$ for each episode. In this setting, our HRL algorithm additionally employs the features of inter-vehicle regions (IVRs) that are subspaces of the action space acquired from our method using domain knowledge. The details of IVRs are in Section \ref{appendix:inter-vehicle_regions} of Appendix.

The subspaces are used in three ways, 1) as additional state features of the IVR that ego-vehicle currently belongs to $c_{t}$, 2) as selectable state features to pay mind to, and 3) as the outline that confines the low-level policy space. The high-level policy $\pi$ chooses one of the candidates $h \in \sH$ whose behavioral mode $b \in \sB$ specifies a set of features of available subspaces to pay mind to, $\sM_{b}$, and the subspace for the outline, $o_{b}$. Thus, a candidate command is defined as $h = \{b, m_{b}, o_{b}\}$ where features of subspace to pay mind to is following $m_{b} \in \sM_{b}$. Table \ref{tab:possible_combinations_for_candidate} in Appendix lists a set of candidates determined by the condition of current state.

The high-level policy $\pi$ at time step $t$ is to select the candidate of maximum state-candidate value $Q_{h}$:
\begin{equation}
\begin{aligned}
    \pi & = \arg\max_{h_{t} \in \sH_{t}} Q(l_{t}, s_{t}, c_{t}, h_{t})\\
        & \simeq \arg\max_{\tilde{h}_{t} \in \tilde{\sH}_{t}} Q(l_{t}, s_{t}, c_{t}, \tilde{h}_{t})\\
        & = \arg\max_{\tilde{h}_{t} \in \tilde{\sH}_{t}} Q(l_{t}, s_{t}, c_{t}, m_{b,t}, b_{t}),
\end{aligned}
\label{high_policy}
\end{equation}
where $\tilde{h}$ is a partial candidate $\{b, m_{b}\}$, and the command is chosen by exhaustive search for all possible partial candidates $\tilde{\sH}_{t}$ at time step $t$. The features of outline region $o_{b}$ are not explicitly indicated in inputs for the value estimation, since $o_{b}$ is either the current IVR or an IVR in mind, and each behavioral mode includes the decision whether to stay in the current IVR or to move on to IVR in mind given the features of both IVRs as inputs $c_{t}$ and $m_{b,t}$. The information on encoding process is described in Section 4 in detail.

In accordance with the high-level policy, our low-level stochastic policy search $\phi$ at time step $t$ generates a two-dimensional goal position $a_{t} \in \sN^{2}$, where $\sN = (0,1)$, given a high-level command $h_{t}$, such that
\begin{equation}
\begin{aligned}
    a_{t} & \sim \phi(l_{t}, s_{t}, c_{t}, h_{t})\\
          & = \phi(l_{t}, s_{t}, c_{t}, b_{t}, m_{b,t}, o_{b,t}).
\end{aligned}
\end{equation}
The normalized goal position $a_{t}$ is interpreted as a position in the local coordinate of the outline region $o_{b,t}$, and our manually designed function 
$\mathcal{T}: \sN^{2} \rightarrow \R^{2}$ transforms the local position $a_{t}$ to the corresponding position in the global coordinate $g_{t}$ given $o_{b,t}$:
\begin{equation}
 g_{t} = \mathcal{T} (a_{t}, o_{b,t}).
\end{equation}
The details of the transformation function $\mathcal{T}$ are in Subsection \ref{appendix:NQC} and \ref{appendix:inter-vehicle_regions} of Appendix. For control of vehicle in reaching the goal $g_{t}$, the outline region $o_{b,t}$ also determines the target heading direction $\psi$.\\
\newline
Given high-level command at every time step, the actual goal position of the agent is determined by $\phi$, thus the high-level state-command value is equal to the state value of low-level policy working in accordance with the command,
\begin{equation}
 Q_{h}(s_{t}, h_{t}) = V_{l}(s_{t}, h_{t}).
\end{equation}
By designing $\phi$ to be learned through actor-critic with the state value estimation, all state-candidate values for our high-level policy can be estimated. When any low-level state value is well estimated, and the low-level policy is locally optimal in a given inter-vehicle region, we can naturally assume that the high-level policy that selects the candidate of maximum state-candidate value as the command is also optimal.

To achieve this global optimum requires nothing but quality learning through actor-critic for action in continuous space, and we adopted proximal policy optimization (PPO) \citep{DBLP:journals/corr/SchulmanWDRK17} to prevent an abrupt decrease in the optimality of learning. The surrogate objective of PPO for our advantage actor-critic given high-level command is
\begin{equation}
\begin{split}
L^{CLIP}_{NEWTRO}(\theta) = \E_{\tau} \sum_{t=t_{0}}^{T-1} min \{w_{t}(\theta) A_{l} (l_{t}, s_{t}, c_{t}, \tilde{h}_{t}, a_{t}),w^{clip}_{t}(\theta) A_{l} (l_{t}, s_{t}, c_{t}, \tilde{h}_{t}, a_{t})\},
\end{split}
\end{equation}
where the clipped ratio is defined as $w^{clip}_{t} = clip (\frac{ \pi_{\theta} (a_{t}|l_{t}, s_{t}, c_{t}, \tilde{h}_{t}) } { pi_{\theta_{old}} (a_{t}|l_{t}, s_{t}, c_{t}, \tilde{h}_{t}) }, 1 - \epsilon, 1 + \epsilon)$ and the trajectory is given for time $t_{0}$ to $T$.\\
\newline
The advantage is equal to the difference between the expected return and the state value given high-level command,
\begin{equation}
 A_{l}(l_{t}, s_{t}, c_{t}, \tilde{h}_{t}, a_{t}) = G_{t} - V_{l}(l_{t}, s_{t}, c_{t}, \tilde{h}_{t}).
\end{equation}
Canonical policy gradient methods \citep{DBLP:conf/nips/SuttonMSM99, DBLP:phd/ndltd/Konda02} learn from recent trajectories for smooth update of policy parameters, and PPO employs a type of estimator for expected returns $G$ introduced in past work \citep{DBLP:journals/ml/Williams92, DBLP:conf/icml/MnihBMGLHSK16}. We expand the estimator as
\begin{equation}
\hat{G_{t}} = r_{t} + \gamma r_{t+1} + \cdots + \gamma^{T-t+1} r_{T-1} r_{t+1} + \gamma^{T-t} V_{max}(l_{T}, s_{T}, c_{T}, \tilde{\sH}_{T}).
\label{expected_return}
\end{equation}
In the last term in the right side of Equation \ref{expected_return}, $V_{max}(l_{T}, s_{T}, c_{T}, \tilde{\sH}_{T})$ is the maximum state-candidate value at time step $T$ acquired from Equation \ref{high_policy} if $T$ is a non-terminal time step, otherwise $0$. Fortunately, this type of expected returns are not tricky to save and batch, since the returns do not require the set of $\tilde{\sH}$ from $t$ to $T-1$ that could include varying number of candidate regions to pay mind to for each behavior through time. In practice, $G_{t}$ is easily implemented by adding $\tilde{\sH}_{T}$ to the batched trajectory at the training time step. \begin{algorithm}
Initialize the parameters of a type of Actor-Critic networks, $\Theta = \{\theta, \vartheta, ... \}$\\
\For{iteration=1,2,...} {
    Choose $h = \{b, m_{b}, o_{b}\} \in \sH$ given $l$, $s$, $c$, $b$, and $m_{b}$ by max-Q policy of parameters $\vartheta$\\
    Choose $a$ given $l$, $s$, $c$, $b$, $m_{b}$, and $o_{b}$, by policy search of parameters $\theta$\\
    Transform $a$ to the goal position $g$ by $\mathcal{T}(a, o_{b})$, where $o_{b}$ defines the local coordinate for $a$\\
    Calculate the target heading direction $\psi$ by using $o_{b}$\\
    Apply $g$ and $\psi$ to the controller of the agent\\
    Save the transition $<l, s, c, b, m_{b}, a, r>$ to the memory\\
    \If{memory size is equal to $T-t_{0}$} {
        Batch transitions, and put the batch and $\tilde{H}_{T}$ as training data\\
        Optimize for policy surrogate objective and critic loss $L$ w.r.t. $\Theta$\\
        Clear memory
    }
}
\caption{SHRL with proximal policy optimization}
\label{SHRL_algorithm}
\end{algorithm}
In brief, the main computation flow of our method is summarized in Algorithm \ref{SHRL_algorithm}, where $\theta$ and $\vartheta$ are parameters of the low-level actor and critic shared in the hierarchy, respectively.
\\

\section{Deep neural network structure and processing} \label{structure}
\begin{figure}[ht]
\begin{center}
\includegraphics[width=1.0\textwidth]{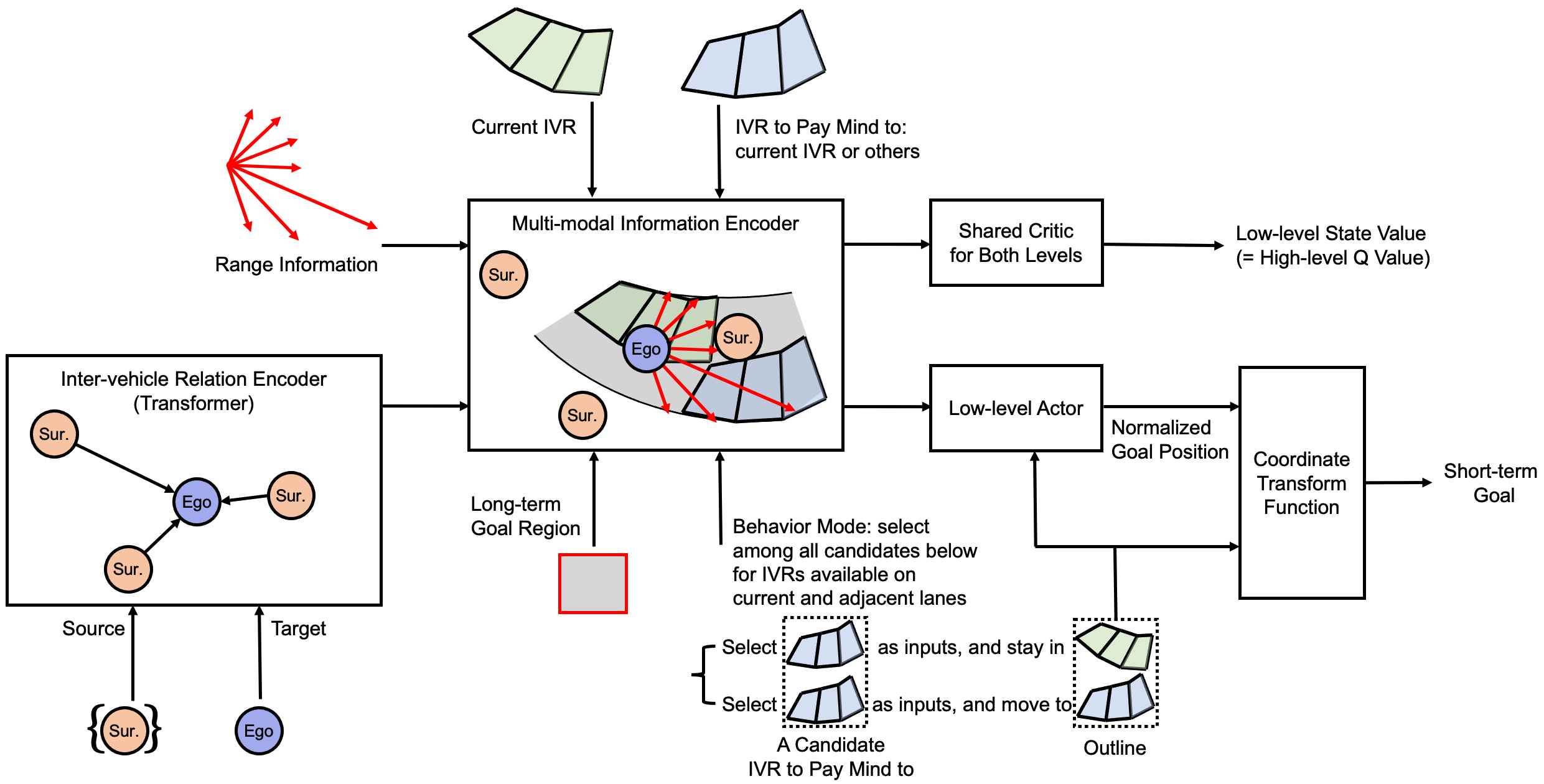}
\end{center}
\label{fig:model_structure}
\caption{Model architecture for our method in autonomous driving}
\end{figure}
The general learning model for AD deals with multi-modal features. In our environment, the model receives three types of inputs: information on vehicles, obstacle positions, and inter-vehicle regions (IVRs). The detailed features are described in Subsection ~\ref{appendix:other_state_features} in Appendix. All types of information include two-dimensional global positions such that the relative information of all feature types is expected to be effectively encoded in our model. Our model consists of four distinctive modules: inter-vehicle relation encoder, multi-modal information encoder, policy networks, and value networks. The overall structure of the modules is described in Figure 1. Actor and critic are implemented with fully-connected layers, and other modules are explained in Subsections below. We practically employed double-critics for stable value learning, which is proposed for deep Q networks \citep{DBLP:conf/aaai/HasseltGS16}.
\subsection{Inter-vehicle relation encoder} \label{ivr_encoder}
Inter-vehicle relation encoder deals with feature information on ego vehicle and surrounding vehicles. To integrate the features of a varying number of surrounding vehicles in respect of ego vehicle, an architecture of graph attention networks, Transformer \citep{mnih2016asynchronous} is used for the encoder. Transformer was originally proposed for natural language processing, and \citet{DBLP:journals/corr/abs-1911-12250} used it in AD to encode the state of vehicles to make high-level decisions for longitudinal control. Transformer consists of encoder layers and decoder layers, and other operations. We used one layer of decoder without encoding layers, where the $source$ inputs receive the features of surrounding vehicles and the $target$ inputs receive the features of ego vehicle. The outputs of inter-vehicle relation encoder are used as one type of the inputs for multi-modal information encoder.
\subsection{Multi-modal information encoder} \label{mmi_encoder}
As seen in Figure 1, multi-modal information encoder integrates the long-term goal $l$, the vehicle encoding $v$, range information $r$, the IVR that the ego belongs to $c$, and the IVR in mind $m$, depending on the behavior mode $b$. The outputs are the encoding of goal, state, and the high-level command used for actor and critic. Each behavioral mode relates these features while interpreting them on the corresponding outline region whose features are received as $c$ or $m$. For behavioral mode, we will present the performance two types of designs: 1) indexed selection for tabular encoding outputs, 2) network attention from the one-hot vector $b$ on hidden layers, $a = f^{attend{1}}(b)$, $h = f^{abstract_{1}}(m)$, and $e = f^{abstract_{2}}(h \odot a)$, which is similarly used in HRL \citep{DBLP:conf/icml/VezhnevetsOSHJS17, earle2018incremental}.
The two designs allow that common features and skills are shared across behavioral modes. The designs were compared in our experiment to find a better structure.
\subsection{Neural elaboration within reinforced outline (NEWTRO)}
The actor plays the role of the low-level policy, generating a goal for the two-dimensional center position, which should be confined to the inside of the outline of the given IVR. For the confinement, we adopt a simple normalization, applying a sigmoid function with the input coefficient $sigmoid(cx)$ as the activation function for the outputs of the actor network. This compares favorably with constraint optimization methods in that it provides faster optimization speed, and that the corners of the ego vehicle mostly stay inside the outline since the goal does not shift closer to the border of the outline unless it is inevitable and values of huge magnitudes are output before the activation. One problem of simply using the output normalization is that both the positions of multi-modal input features that construct the encoding along with behavior and the featuring positions of the outline IVR are specified with respect to the global coordinate frame, while the outputs of the actor are normalized between $0$ and $1$. Thus, we devise an additional normalization technique for the multi-modal information encoding $e$ to train the actor to be invariant of the varying size of the outline. Before being used as inputs for the actor network, $e$ is simply copied and divided by the length and sampled widths of the outline. An IVR is given as the positions of the left and right sides sampled for $N$ times along the progressing direction of vehicles. The length of the outline along the progressing direction, $l_{o}$, and the widths of the region, $\{w_{o,1}, w_{o,2}, \cdot, w_{o,N-1}, w_{o,N}\}$, which are the distances between the two sides of each sample, are defined by the method in Subsection \ref{appendix:lane_approximation} and \ref{appendix:inter-vehicle_regions} of Appendix. Thus, $e$ of $d$ dimensions is copied $N+1$ times, divided by these length and widths, and concatenated so that $e$ turns to $\{e/l_{o}, e/w_{o,1}, e/w_{o,2}, \cdot, e/w_{o,N-1}, e/w_{o,N}\}$.
Using the copied encodings of $N+1$ dimension normalized by the outline that is selected by the reinforced high-level policy, the local sub-policy network elaborates the outline by generating a goal position.
\\
\section{Experimental results}
\subsection{Experiment setting}
In our environment, agents appear at random lanes at the left end of the road at random time steps, and an agent ends the episode when it reaches the goal region, which is the right end of the road, or collides with either other vehicles or lane shoulders. When a collision occurs, vehicles stay stuck on the road for a fixed delay time and then disappear from the road. The reward given to the agent differs depending on the significance of objectives. Big reward or penalty is given at the end of the episode while small rewards or penalties is given during the episode. The details of rewards are described in the table in Section \ref{appendix:other_state_features} of Appendix. Vehicle agents are equipped with Stanley controller \citep{hoffmann2007autonomous} for lateral control and hierarchical PID controllers for longitudinal control, and simulated by a kinematic bicycle model \citep{DBLP:conf/ivs/KongPSB15}. We prepared environments with various types of roads, which are in Figure 4 of Appendix. To average the optimization of multi-agents and simplify the comparison of performance for different methods in Figure 2, we adopted a joint training method of \citet{mnih2016asynchronous}, which is originally for single-agent in multi-environments.

\subsection{Evaluation}
\begin{figure}[h]
\begin{center}
\includegraphics[width=0.85\textwidth]{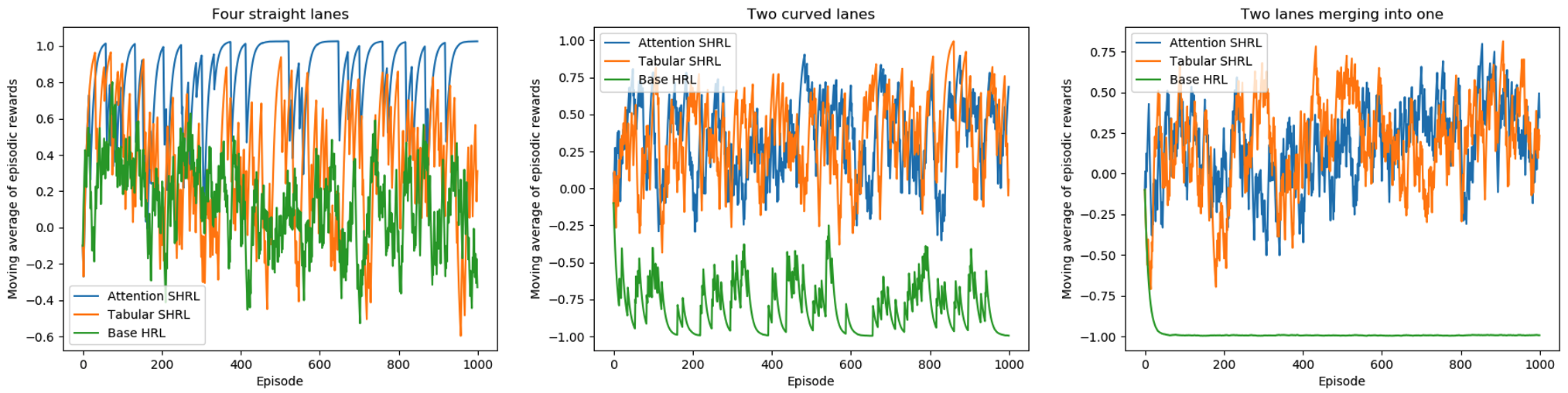}
\end{center}
\label{fig:episodic_rewards}
\caption{Training evolution of our methods and a baseline on three types of roads}
\end{figure}
We compare SHRL of two state-command encoding structures to a baseline HRL method with domain knowledge about the definition of sub-goals. Figure 2 shows the performance of agents on the roads of four straight lanes (left), two curved lanes (center), and two lanes merging into one (right). During the joint training of all vehicle agents using each method, we plotted moving average of the sum of returns for each episode with initiation to $0$ and $alpha = 0.1$.

The total episode rewards are close to $1$ when an agent successfully learns the task including goal reaching, and when an agent fails and have an accident, $-1$. In comparison with the baseline HRL, the attention-based methods for the behavior mode showed decent performance from early episodes for all types of roads taking advantage of the outline. When the agent collides with surrounding vehicles, the episodic rewards drops for a moment, and steadily increase through learning multi-agent interactions. We also tested the structural design using separate modules for each mode, but it was not as effective as the two methods. The baseline method showed fluctuating performance for straight lanes, low performance for the curved lanes, and failed to learn in the merging lanes since safe short-term goals were hardly constructed as the chance of collision increases.

\subsection{Visualization}
\begin{figure}[h]
\begin{center}
\includegraphics[width=0.4\textwidth]{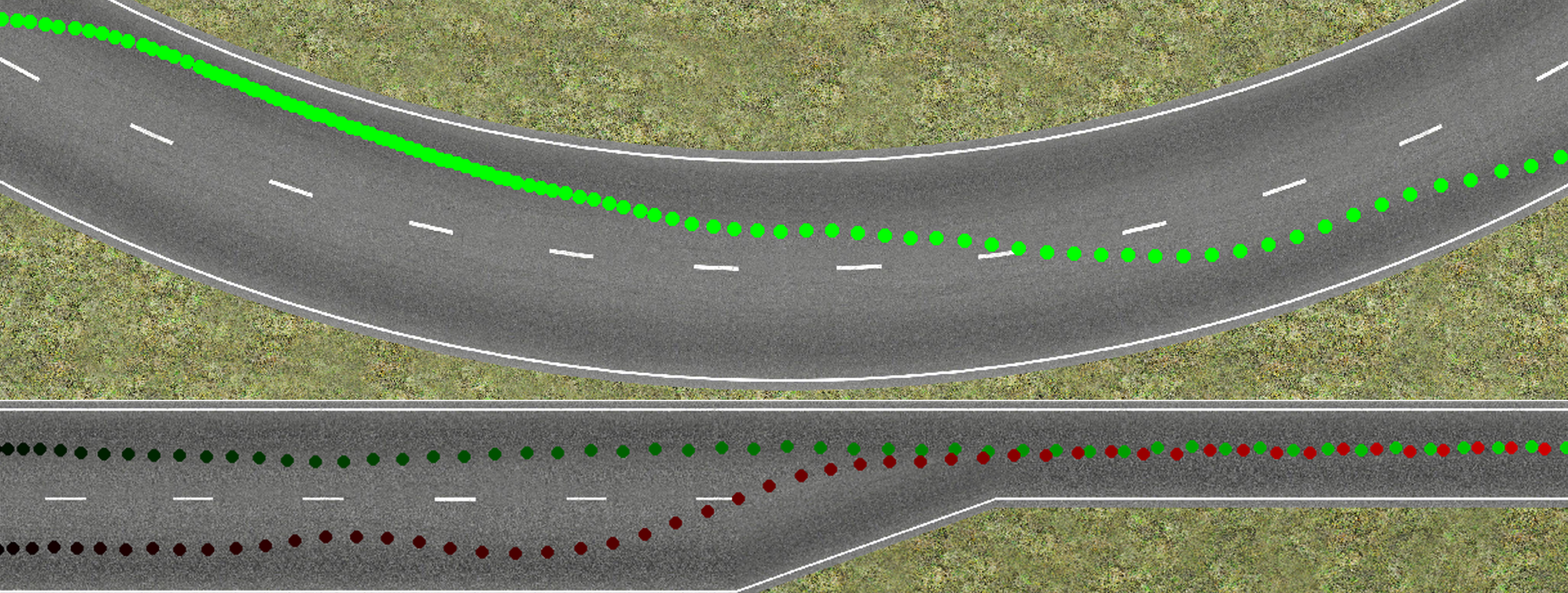}
\end{center}
\label{fig:traj}
\caption{Trajectories of vehicles on two curved lanes}
\end{figure}
We qualitatively examined the traveling trajectory of vehicle agents on curved two lanes in our method against that of the baseline HRL method. 
General human tactics for the agent is to take the shortest possible path, which could be a straight line in the optimal environment. The agent should consider collision avoidance, travel time reduction, and compliance with speed limits.
In our experiment, two vehicle agents start to drive from the left end of each lane almost simultaneously with slight randomness. When an inter-vehicle region on the shortest possible path is available, then the agent will move on to occupy the region. Figure 3 shows agents using our method. The agent on two curved lane tried to take the inner lane at first, and then the outer lane in the end. The red agent on merging lane wait until the green agents pass the lane in the middle, and follow the green agent. Agents using the baseline HRL, with frequent accidents, and seemingly not following the optimal path even when they complete the journey.
\section{Discussion and future work}
We presented hierarchical policies that can utilize space decomposition of state and policy by selecting a subspace at the high level and learning low-level policy with the given subspace. Although the hierarchical policies are from different base value definitions that are high-level state-action value and low-level state value given command, we integrated them as the same network structure learned by single-level training. Thus, our method can be utilized as a part of a nested spatio-temporal hierarchy of reinforcement learning where inner spatial hierarchy is implemented with our method and outer temporal hierarchy is designed with canonical HRL methods.

For autonomous driving, our experiments showed that learning the short-term planning using domain knowledge about collision-free space and its decomposition is extremely efficient. We expect that the use of domain knowledge in deep reinforcement learning with integrated value in the hierarchy, which is seemingly close to the basis of human cognitive thinking and proven to be efficient can be applied to complex levels of reinforcement learning: model-based reinforcement learning with predictive control or self-interested multi-agent reinforcement learning in game-theoretic situations.

The limitation of our method is that a separate process for spatial decomposition is required, and that the performance highly depends on the decomposition ability. Effective automatic spatial decomposition methods through unsupervised learning or reinforcement learning will be able to generalize our HRL algorithm to different tasks.

\bibliographystyle{unsrtnat}
\bibliography{bibs/pg, bibs/actorcritic, bibs/hrl, bibs/ad, bibs/goal_rl, bibs/adrl, bibs/adhrl, bibs/dqn, bibs/continuousaction, bibs/gnns, bibs/her, bibs/simmodelandcontrol, bibs/stateaggregationrl, bibs/stateattention, bibs/trpo, bibs/attentionhrl}

\begin{thebibliography}{32}
\providecommand{\natexlab}[1]{#1}
\providecommand{\url}[1]{\texttt{#1}}
\expandafter\ifx\csname urlstyle\endcsname\relax
  \providecommand{\doi}[1]{doi: #1}\else
  \providecommand{\doi}{doi: \begingroup \urlstyle{rm}\Url}\fi

\bibitem[Parr and Russell(1998)]{parr1998reinforcement}
Ronald Parr and Stuart Russell.
\newblock Reinforcement learning with hierarchies of machines.
\newblock \emph{Advances in neural information processing systems}, pages
  1043--1049, 1998.

\bibitem[Sutton et~al.(1999{\natexlab{a}})Sutton, Precup, and
  Singh]{sutton1999between}
Richard~S Sutton, Doina Precup, and Satinder Singh.
\newblock Between mdps and semi-mdps: A framework for temporal abstraction in
  reinforcement learning.
\newblock \emph{Artificial intelligence}, 112\penalty0 (1-2):\penalty0
  181--211, 1999{\natexlab{a}}.

\bibitem[Dayan and Hinton(1992)]{DBLP:conf/nips/DayanH92}
Peter Dayan and Geoffrey~E. Hinton.
\newblock Feudal reinforcement learning.
\newblock In Stephen~Jose Hanson, Jack~D. Cowan, and C.~Lee Giles, editors,
  \emph{Advances in Neural Information Processing Systems 5, {[NIPS}
  Conference, Denver, Colorado, USA, November 30 - December 3, 1992]}, pages
  271--278. Morgan Kaufmann, 1992.
\newblock URL
  \url{http://papers.nips.cc/paper/714-feudal-reinforcement-learning}.

\bibitem[Singh et~al.(1994)Singh, Jaakkola, and
  Jordan]{DBLP:conf/nips/SinghJJ94}
Satinder~P. Singh, Tommi~S. Jaakkola, and Michael~I. Jordan.
\newblock Reinforcement learning with soft state aggregation.
\newblock In Gerald Tesauro, David~S. Touretzky, and Todd~K. Leen, editors,
  \emph{Advances in Neural Information Processing Systems 7, {[NIPS}
  Conference, Denver, Colorado, USA, 1994]}, pages 361--368. {MIT} Press, 1994.
\newblock URL
  \url{http://papers.nips.cc/paper/981-reinforcement-learning-with-soft-state-aggregation}.

\bibitem[Dietterich(2000)]{dietterich2000hierarchical}
Thomas~G Dietterich.
\newblock Hierarchical reinforcement learning with the maxq value function
  decomposition.
\newblock \emph{Journal of artificial intelligence research}, 13:\penalty0
  227--303, 2000.

\bibitem[Kulkarni et~al.(2016)Kulkarni, Narasimhan, Saeedi, and
  Tenenbaum]{NIPS2016_f442d33f}
Tejas~D Kulkarni, Karthik Narasimhan, Ardavan Saeedi, and Josh Tenenbaum.
\newblock Hierarchical deep reinforcement learning: Integrating temporal
  abstraction and intrinsic motivation.
\newblock In D.~Lee, M.~Sugiyama, U.~Luxburg, I.~Guyon, and R.~Garnett,
  editors, \emph{Advances in Neural Information Processing Systems}, volume~29.
  Curran Associates, Inc., 2016.
\newblock URL
  \url{https://proceedings.neurips.cc/paper/2016/file/f442d33fa06832082290ad8544a8da27-Paper.pdf}.

\bibitem[Nachum et~al.(2018)Nachum, Gu, Lee, and
  Levine]{DBLP:conf/nips/NachumGLL18}
Ofir Nachum, Shixiang Gu, Honglak Lee, and Sergey Levine.
\newblock Data-efficient hierarchical reinforcement learning.
\newblock In Samy Bengio, Hanna~M. Wallach, Hugo Larochelle, Kristen Grauman,
  Nicol{\`{o}} Cesa{-}Bianchi, and Roman Garnett, editors, \emph{Advances in
  Neural Information Processing Systems 31: Annual Conference on Neural
  Information Processing Systems 2018, NeurIPS 2018, December 3-8, 2018,
  Montr{\'{e}}al, Canada}, pages 3307--3317, 2018.
\newblock URL
  \url{https://proceedings.neurips.cc/paper/2018/hash/e6384711491713d29bc63fc5eeb5ba4f-Abstract.html}.

\bibitem[Bacon et~al.(2017)Bacon, Harb, and Precup]{DBLP:conf/aaai/BaconHP17}
Pierre{-}Luc Bacon, Jean Harb, and Doina Precup.
\newblock The option-critic architecture.
\newblock In Satinder~P. Singh and Shaul Markovitch, editors, \emph{Proceedings
  of the Thirty-First {AAAI} Conference on Artificial Intelligence, February
  4-9, 2017, San Francisco, California, {USA}}, pages 1726--1734. {AAAI} Press,
  2017.
\newblock URL \url{http://aaai.org/ocs/index.php/AAAI/AAAI17/paper/view/14858}.

\bibitem[Haarnoja et~al.(2018)Haarnoja, Hartikainen, Abbeel, and
  Levine]{DBLP:conf/icml/HaarnojaHAL18}
Tuomas Haarnoja, Kristian Hartikainen, Pieter Abbeel, and Sergey Levine.
\newblock Latent space policies for hierarchical reinforcement learning.
\newblock In Jennifer~G. Dy and Andreas Krause, editors, \emph{Proceedings of
  the 35th International Conference on Machine Learning, {ICML} 2018,
  Stockholmsm{\"{a}}ssan, Stockholm, Sweden, July 10-15, 2018}, volume~80 of
  \emph{Proceedings of Machine Learning Research}, pages 1846--1855. {PMLR},
  2018.
\newblock URL \url{http://proceedings.mlr.press/v80/haarnoja18a.html}.

\bibitem[Li et~al.(2020)Li, Florensa, Clavera, and
  Abbeel]{DBLP:conf/iclr/LiFCA20}
Alexander~C. Li, Carlos Florensa, Ignasi Clavera, and Pieter Abbeel.
\newblock Sub-policy adaptation for hierarchical reinforcement learning.
\newblock In \emph{8th International Conference on Learning Representations,
  {ICLR} 2020, Addis Ababa, Ethiopia, April 26-30, 2020}. OpenReview.net, 2020.
\newblock URL \url{https://openreview.net/forum?id=ByeWogStDS}.

\bibitem[Paxton et~al.(2017)Paxton, Raman, Hager, and
  Kobilarov]{DBLP:conf/iros/PaxtonRHK17}
Chris Paxton, Vasumathi Raman, Gregory~D. Hager, and Marin Kobilarov.
\newblock Combining neural networks and tree search for task and motion
  planning in challenging environments.
\newblock In \emph{2017 {IEEE/RSJ} International Conference on Intelligent
  Robots and Systems, {IROS} 2017, Vancouver, BC, Canada, September 24-28,
  2017}, pages 6059--6066. {IEEE}, 2017.
\newblock \doi{10.1109/IROS.2017.8206505}.
\newblock URL \url{https://doi.org/10.1109/IROS.2017.8206505}.

\bibitem[Chen et~al.(2019)Chen, Dong, Palanisamy, Mudalige, Muelling, and
  Dolan]{DBLP:conf/iros/ChenDPMMD19}
Yilun Chen, Chiyu Dong, Praveen Palanisamy, Priyantha Mudalige, Katharina
  Muelling, and John~M. Dolan.
\newblock Attention-based hierarchical deep reinforcement learning for lane
  change behaviors in autonomous driving.
\newblock In \emph{2019 {IEEE/RSJ} International Conference on Intelligent
  Robots and Systems, {IROS} 2019, Macau, SAR, China, November 3-8, 2019},
  pages 3697--3703. {IEEE}, 2019.
\newblock \doi{10.1109/IROS40897.2019.8968565}.
\newblock URL \url{https://doi.org/10.1109/IROS40897.2019.8968565}.

\bibitem[Rezaee et~al.(2019)Rezaee, Yadmellat, Nosrati, Abolfathi, Elmahgiubi,
  and Luo]{DBLP:conf/itsc/RezaeeYNAEL19}
Kasra Rezaee, Peyman Yadmellat, Masoud~S. Nosrati, Elmira~Amirloo Abolfathi,
  Mohammed Elmahgiubi, and Jun Luo.
\newblock Multi-lane cruising using hierarchical planning and reinforcement
  learning.
\newblock In \emph{2019 {IEEE} Intelligent Transportation Systems Conference,
  {ITSC} 2019, Auckland, New Zealand, October 27-30, 2019}, pages 1800--1806.
  {IEEE}, 2019.
\newblock \doi{10.1109/ITSC.2019.8916928}.
\newblock URL \url{https://doi.org/10.1109/ITSC.2019.8916928}.

\bibitem[Schaul et~al.(2015)Schaul, Horgan, Gregor, and
  Silver]{DBLP:conf/icml/SchaulHGS15}
Tom Schaul, Daniel Horgan, Karol Gregor, and David Silver.
\newblock Universal value function approximators.
\newblock In Francis~R. Bach and David~M. Blei, editors, \emph{Proceedings of
  the 32nd International Conference on Machine Learning, {ICML} 2015, Lille,
  France, 6-11 July 2015}, volume~37 of \emph{{JMLR} Workshop and Conference
  Proceedings}, pages 1312--1320. JMLR.org, 2015.
\newblock URL \url{http://proceedings.mlr.press/v37/schaul15.html}.

\bibitem[Sutton et~al.(1999{\natexlab{b}})Sutton, McAllester, Singh, and
  Mansour]{DBLP:conf/nips/SuttonMSM99}
Richard~S. Sutton, David~A. McAllester, Satinder~P. Singh, and Yishay Mansour.
\newblock Policy gradient methods for reinforcement learning with function
  approximation.
\newblock In Sara~A. Solla, Todd~K. Leen, and Klaus{-}Robert M{\"{u}}ller,
  editors, \emph{Advances in Neural Information Processing Systems 12, {[NIPS}
  Conference, Denver, Colorado, USA, November 29 - December 4, 1999]}, pages
  1057--1063. The {MIT} Press, 1999{\natexlab{b}}.
\newblock URL
  \url{http://papers.nips.cc/paper/1713-policy-gradient-methods-for-reinforcement-learning-with-function-approximation}.

\bibitem[Konda(2002)]{DBLP:phd/ndltd/Konda02}
Vijaymohan Konda.
\newblock \emph{Actor-critic algorithms}.
\newblock PhD thesis, Massachusetts Institute of Technology, Cambridge, MA,
  {USA}, 2002.
\newblock URL \url{http://hdl.handle.net/1721.1/8120}.

\bibitem[Schulman et~al.(2015)Schulman, Moritz, Levine, Jordan, and
  Abbeel]{schulman2015high}
John Schulman, Philipp Moritz, Sergey Levine, Michael Jordan, and Pieter
  Abbeel.
\newblock High-dimensional continuous control using generalized advantage
  estimation.
\newblock \emph{arXiv preprint arXiv:1506.02438}, 2015.

\bibitem[Chen et~al.(2017)Chen, Zhai, Lu, Gong, and
  Wang]{DBLP:conf/ivs/ChenZLGW17}
Xin Chen, Yong Zhai, Chao Lu, Jianwei Gong, and Gang Wang.
\newblock A learning model for personalized adaptive cruise control.
\newblock In \emph{{IEEE} Intelligent Vehicles Symposium, {IV} 2017, Los
  Angeles, CA, USA, June 11-14, 2017}, pages 379--384. {IEEE}, 2017.
\newblock \doi{10.1109/IVS.2017.7995748}.
\newblock URL \url{https://doi.org/10.1109/IVS.2017.7995748}.

\bibitem[Zhao et~al.(2017)Zhao, Xia, and Zhang]{DBLP:journals/cim/ZhaoXZ17}
Dongbin Zhao, Zhongpu Xia, and Qichao Zhang.
\newblock Model-free optimal control based intelligent cruise control with
  hardware-in-the-loop demonstration [research frontier].
\newblock \emph{{IEEE} Comput. Intell. Mag.}, 12\penalty0 (2):\penalty0 56--69,
  2017.
\newblock \doi{10.1109/MCI.2017.2670463}.
\newblock URL \url{https://doi.org/10.1109/MCI.2017.2670463}.

\bibitem[Kendall et~al.(2019)Kendall, Hawke, Janz, Mazur, Reda, Allen, Lam,
  Bewley, and Shah]{DBLP:conf/icra/KendallHJMRALBS19}
Alex Kendall, Jeffrey Hawke, David Janz, Przemyslaw Mazur, Daniele Reda,
  John{-}Mark Allen, Vinh{-}Dieu Lam, Alex Bewley, and Amar Shah.
\newblock Learning to drive in a day.
\newblock In \emph{International Conference on Robotics and Automation, {ICRA}
  2019, Montreal, QC, Canada, May 20-24, 2019}, pages 8248--8254. {IEEE}, 2019.
\newblock \doi{10.1109/ICRA.2019.8793742}.
\newblock URL \url{https://doi.org/10.1109/ICRA.2019.8793742}.

\bibitem[Wang et~al.(2018)Wang, Chan, and
  de~La~Fortelle]{DBLP:conf/ivs/WangCF18}
Pin Wang, Ching{-}Yao Chan, and Arnaud de~La~Fortelle.
\newblock A reinforcement learning based approach for automated lane change
  maneuvers.
\newblock In \emph{2018 {IEEE} Intelligent Vehicles Symposium, {IV} 2018,
  Changshu, Suzhou, China, June 26-30, 2018}, pages 1379--1384. {IEEE}, 2018.
\newblock \doi{10.1109/IVS.2018.8500556}.
\newblock URL \url{https://doi.org/10.1109/IVS.2018.8500556}.

\bibitem[Wang and Chan(2017)]{DBLP:conf/itsc/WangC17}
Pin Wang and Ching{-}Yao Chan.
\newblock Formulation of deep reinforcement learning architecture toward
  autonomous driving for on-ramp merge.
\newblock In \emph{20th {IEEE} International Conference on Intelligent
  Transportation Systems, {ITSC} 2017, Yokohama, Japan, October 16-19, 2017},
  pages 1--6. {IEEE}, 2017.
\newblock \doi{10.1109/ITSC.2017.8317735}.
\newblock URL \url{https://doi.org/10.1109/ITSC.2017.8317735}.

\bibitem[Schulman et~al.(2017)Schulman, Wolski, Dhariwal, Radford, and
  Klimov]{DBLP:journals/corr/SchulmanWDRK17}
John Schulman, Filip Wolski, Prafulla Dhariwal, Alec Radford, and Oleg Klimov.
\newblock Proximal policy optimization algorithms.
\newblock \emph{CoRR}, abs/1707.06347, 2017.
\newblock URL \url{http://arxiv.org/abs/1707.06347}.

\bibitem[Williams(1992)]{DBLP:journals/ml/Williams92}
Ronald~J. Williams.
\newblock Simple statistical gradient-following algorithms for connectionist
  reinforcement learning.
\newblock \emph{Mach. Learn.}, 8:\penalty0 229--256, 1992.
\newblock \doi{10.1007/BF00992696}.
\newblock URL \url{https://doi.org/10.1007/BF00992696}.

\bibitem[Mnih et~al.(2016{\natexlab{a}})Mnih, Badia, Mirza, Graves, Lillicrap,
  Harley, Silver, and Kavukcuoglu]{DBLP:conf/icml/MnihBMGLHSK16}
Volodymyr Mnih, Adri{\`{a}}~Puigdom{\`{e}}nech Badia, Mehdi Mirza, Alex Graves,
  Timothy~P. Lillicrap, Tim Harley, David Silver, and Koray Kavukcuoglu.
\newblock Asynchronous methods for deep reinforcement learning.
\newblock In Maria{-}Florina Balcan and Kilian~Q. Weinberger, editors,
  \emph{Proceedings of the 33nd International Conference on Machine Learning,
  {ICML} 2016, New York City, NY, USA, June 19-24, 2016}, volume~48 of
  \emph{{JMLR} Workshop and Conference Proceedings}, pages 1928--1937.
  JMLR.org, 2016{\natexlab{a}}.
\newblock URL \url{http://proceedings.mlr.press/v48/mniha16.html}.

\bibitem[van Hasselt et~al.(2016)van Hasselt, Guez, and
  Silver]{DBLP:conf/aaai/HasseltGS16}
Hado van Hasselt, Arthur Guez, and David Silver.
\newblock Deep reinforcement learning with double q-learning.
\newblock In Dale Schuurmans and Michael~P. Wellman, editors, \emph{Proceedings
  of the Thirtieth {AAAI} Conference on Artificial Intelligence, February
  12-17, 2016, Phoenix, Arizona, {USA}}, pages 2094--2100. {AAAI} Press, 2016.
\newblock URL
  \url{http://www.aaai.org/ocs/index.php/AAAI/AAAI16/paper/view/12389}.

\bibitem[Mnih et~al.(2016{\natexlab{b}})Mnih, Badia, Mirza, Graves, Lillicrap,
  Harley, Silver, and Kavukcuoglu]{mnih2016asynchronous}
Volodymyr Mnih, Adria~Puigdomenech Badia, Mehdi Mirza, Alex Graves, Timothy
  Lillicrap, Tim Harley, David Silver, and Koray Kavukcuoglu.
\newblock Asynchronous methods for deep reinforcement learning.
\newblock In \emph{International conference on machine learning}, pages
  1928--1937. PMLR, 2016{\natexlab{b}}.

\bibitem[Leurent and Mercat(2019)]{DBLP:journals/corr/abs-1911-12250}
Edouard Leurent and Jean Mercat.
\newblock Social attention for autonomous decision-making in dense traffic.
\newblock \emph{CoRR}, abs/1911.12250, 2019.
\newblock URL \url{http://arxiv.org/abs/1911.12250}.

\bibitem[Vezhnevets et~al.(2017)Vezhnevets, Osindero, Schaul, Heess, Jaderberg,
  Silver, and Kavukcuoglu]{DBLP:conf/icml/VezhnevetsOSHJS17}
Alexander~Sasha Vezhnevets, Simon Osindero, Tom Schaul, Nicolas Heess, Max
  Jaderberg, David Silver, and Koray Kavukcuoglu.
\newblock Feudal networks for hierarchical reinforcement learning.
\newblock In Doina Precup and Yee~Whye Teh, editors, \emph{Proceedings of the
  34th International Conference on Machine Learning, {ICML} 2017, Sydney, NSW,
  Australia, 6-11 August 2017}, volume~70 of \emph{Proceedings of Machine
  Learning Research}, pages 3540--3549. {PMLR}, 2017.
\newblock URL \url{http://proceedings.mlr.press/v70/vezhnevets17a.html}.

\bibitem[Earle et~al.(2018)Earle, Saxe, and Rosman]{earle2018incremental}
Adam~C Earle, Andrew~M Saxe, and Benjamin Rosman.
\newblock Incremental hierarchical reinforcement learning with multitask lmdps.
\newblock 2018.

\bibitem[Hoffmann et~al.(2007)Hoffmann, Tomlin, Montemerlo, and
  Thrun]{hoffmann2007autonomous}
Gabriel~M Hoffmann, Claire~J Tomlin, Michael Montemerlo, and Sebastian Thrun.
\newblock Autonomous automobile trajectory tracking for off-road driving:
  Controller design, experimental validation and racing.
\newblock In \emph{2007 American Control Conference}, pages 2296--2301. IEEE,
  2007.

\bibitem[Kong et~al.(2015)Kong, Pfeiffer, Schildbach, and
  Borrelli]{DBLP:conf/ivs/KongPSB15}
Jason Kong, Mark Pfeiffer, Georg Schildbach, and Francesco Borrelli.
\newblock Kinematic and dynamic vehicle models for autonomous driving control
  design.
\newblock In \emph{2015 {IEEE} Intelligent Vehicles Symposium, {IV} 2015,
  Seoul, South Korea, June 28 - July 1, 2015}, pages 1094--1099. {IEEE}, 2015.
\newblock \doi{10.1109/IVS.2015.7225830}.
\newblock URL \url{https://doi.org/10.1109/IVS.2015.7225830}.

\end{thebibliography}

\appendix
\section{Appendix}

\subsection{Roads and lane approximation} \label{appendix:lane_approximation}
\begin{figure}[ht]
\begin{center}
\includegraphics[width=0.8\textwidth]{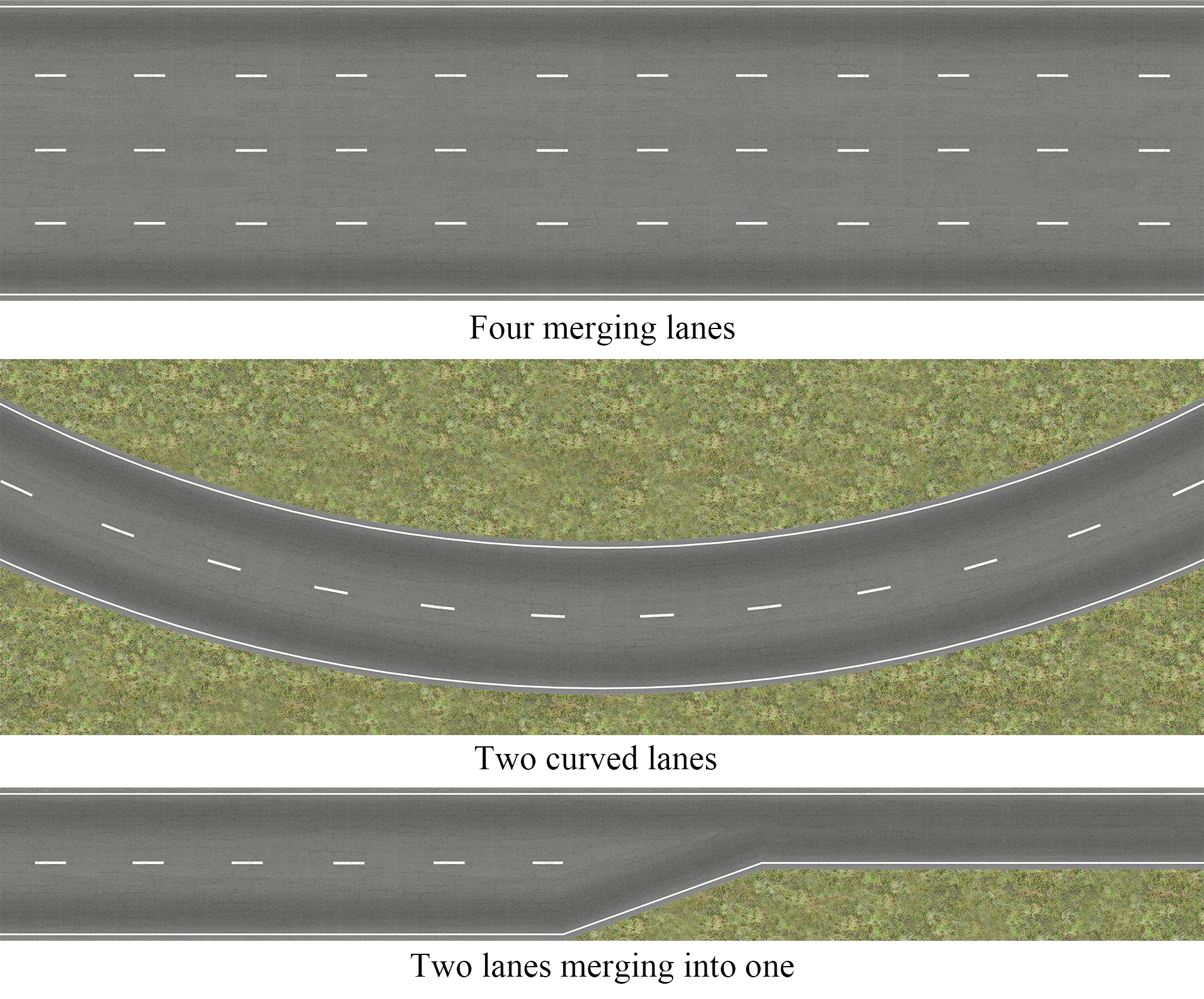}
\end{center}
\label{fig:roads}
\caption{Three types of roads for our highway driving environment}
\end{figure}
Our highway driving environments have one of three types of roads: four straight lanes, two curved lanes, and two lanes merging into one, which are in Figure 4. To handle the information on these various shapes of lanes, the agent uses the sampled points from each lane. For each lane, positions of the left and right sides of the lane are sampled along the lane such that a concatenation of convex quadrilaterals can be created. For each quadrilateral, information on adjacency to another quadrilateral in bordering lanes is given as well.
The quadrilaterals approximating a lane are used to define inter-vehicle regions (IVRs). When two positions at the rear end $(q_{r}, v_{r})$ and the front end $(q_{f}, v_{f})$ are given for a lane $l$, the distance between the two $D(l, (q_{r}, v_{r}), (q_{f}, v_{f}))$ is defined by the sum of distances, $(1-v_{r}) d(Q(l, q_{r})) + \sum_{i=q_{r}+1}^{q_{f}-1} d(Q(l,i)) + v_{f} d(Q(l,q_{f}))$, where $Q(l,q)$ is the quadrilateral search function and $d(Q)$ is the Euclidean distance between the center positions of the rear end $(0.5, 0)$ and the front end $(0.5, 1)$ of the given quadrilateral.
\subsection{Normalized quadrilateral coordinate} \label{appendix:NQC}
\begin{figure}[ht]
\begin{center}
\includegraphics[width=0.8\textwidth]{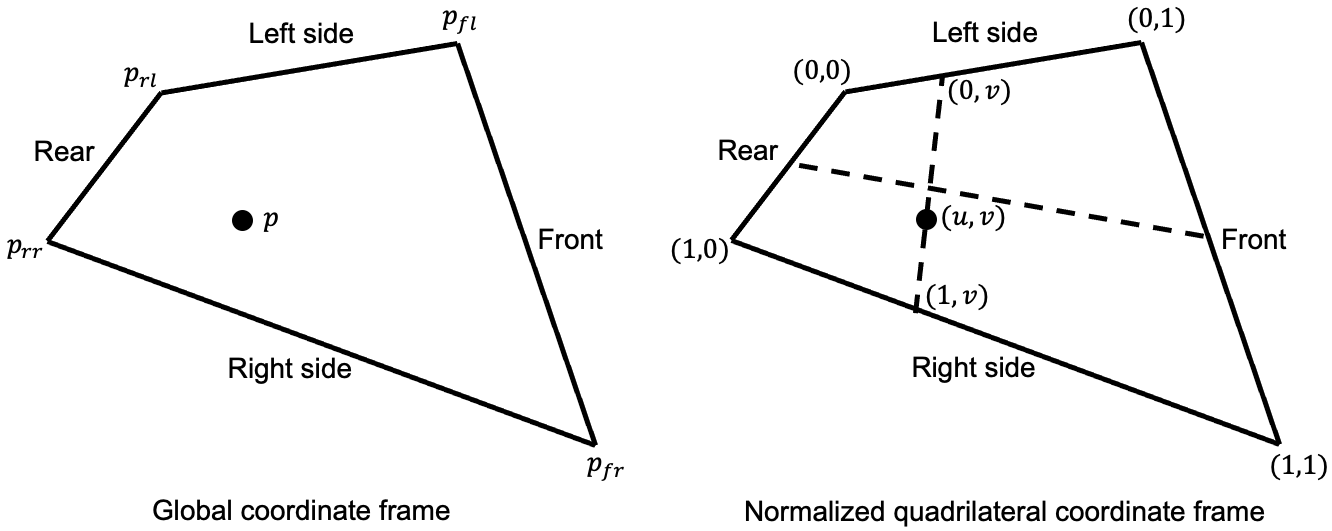}
\end{center}
\label{fig:nqc}
\caption{A position in a quadrilateral represented in normalized quadrilateral coordinate frame}
\end{figure}
We are given four positions of a quadrilateral in a lane with respect to global coordinate or ego-centric coordinate, $(x_{fl}, y_{fl})$, $(x_{fr}, y_{fr})$, $(x_{rl}, y_{rl})$, $(x_{rr}, y_{rr})$, which are positions of front-left, front-right, rear-left, and rear-right. The goal of our transformation is two-fold: First, a position inside the quadrilateral $(x, y)$ should be transformed to the normalized position $(u, v)$ of the given quadrilateral so that $(x_{fl}, y_{fl})$, $(x_{fr}, y_{fr})$, $(x_{rl}, y_{rl})$, and $(x_{rr}, y_{rr})$ correspond to $(1,0)$, $(1,1)$, $(0,0)$, and $(0,1)$. 
Second, $(x, y)$ should be the interpolation of positions on both left and right sides that divide the length of each side in the same proportion $v$ along the progression. Thus, we designed the transformation so that $(x, y)$ is represented as the interpolation of $(x_{vl}, y_{vl})$ and $(x_{vr}, y_{vr})$ with ratio $u$, which are the interpolation of $(x_{fl}, y_{fl})$ and $(x_{rl}, y_{rl})$ with ratio $v$ and the interpolation of $(x_{fr}, y_{fr})$ and $(x_{rr}, y_{rr})$ with ratio $v$, respectively. Figure 5 shows the positions on an arbitrary quadrilateral represented in both global coordinate frame and normalized quadrilateral coordinate frame. The transformation from normalized position $(u, v)$ to global position $(x,y)$ is easily done by the two steps of interpolation, while transformation from $(x,y)$ to $(u, v)$ is performed by solving the following equations. For $x$ axis,
\begin{equation}
\begin{aligned}
   x_{vl} &= x_{rl} + v (x_{fl} - x_{rl})\\
          &= x_{rl} + v d_{l},
\end{aligned}
\end{equation}
\begin{equation}
\begin{aligned}
   x_{vr} &= x_{rr} + v (x_{fr} - x_{rr})\\
          &= x_{rr} + v d_{r},
\end{aligned}
\end{equation}
\begin{equation}
\begin{aligned}
   u &= \frac{(x - x_{vl})}{(x_{vr} - x_{vl})}\\
     &= \frac{x - (x_{rl} + v d_{l})}{(x_{rr} + v d_{r}) - (x_{rl} + v d_{l})}\\
     &= \frac{(x - x_{rl}) - v d_{l}}{v (d_{r} - d_{l}) + (x_{rr} - x_{rl})}\\
     &= \frac{e - v d_{l}}{f + v g},
\end{aligned}
\end{equation}
where $d_{l} = x_{fl} - x_{rl}$, $e = x - x_{rl}$, $f = x_{rr} - x_{rl}$, and $g = (x_{fr} - x_{rr}) - (x_{fl} - x_{rl})$.

In the same way for $y$ axis,
\begin{equation}
   u = \frac{i - v h_{l}}{j + v k},
\end{equation}
where $h_{l} = y_{fl} - y_{rl}$, $i = y - y_{rl}$, $j = y_{rr} - y_{rl}$, and $k = (y_{fr} - y_{rr}) - (y_{fl} - y_{rl})$.

Now a quadratic equation about $v$ is formed by combining the equations about $u$ on both axes.
\begin{equation}
\frac{e - v d_{l}}{f + v g} = \frac{i - v h_{l}}{j + v k}
\label{v_same_for_x_y}
\end{equation}
The Equation \ref{v_same_for_x_y} can be developed to
\begin{equation}
(d_{l} k - h_{l} g) v^{2} + (g i + d_{l} j - h_{l} f - e k) v + (e j - f i) = 0,
\label{v_quadratic_eq}
\end{equation}
which is a quadratic equation for $v$, and this can be substituted to
\begin{equation}
a v^{2} + b v + c = 0,
\label{general_quadratic_eq}
\end{equation}
where $a = d_{l} k - h_{l} g$, $b = (g i + d_{l} j - h_{l} f - e k)$, and $c = e j - f i$.

Since $(x, y)$ is inside the quadrilateral, $u$ and $v$ are between $0$ and $1$. Thus, the solution to Equation \ref{general_quadratic_eq} satisfying the condition is $v = \frac{-b + \sqrt{b^{2} - 4 a c}}{2 a c}$ if $a > 0$,
$v = \frac{-b - \sqrt{b^{2} - 4 a c}}{2 a c}$ if $a < 0$,
otherwise, $v = -c/b$. Then, $u = \frac{e - v d_{l}}{f + v g}$ if $f + v g \neq 0$, otherwise, $\frac{i - v h_{l}}{j + v k}$.
\\
\subsection{Hashing quadrilaterals}
\begin{figure}[h]
\begin{center}
\includegraphics[width=0.5\textwidth]{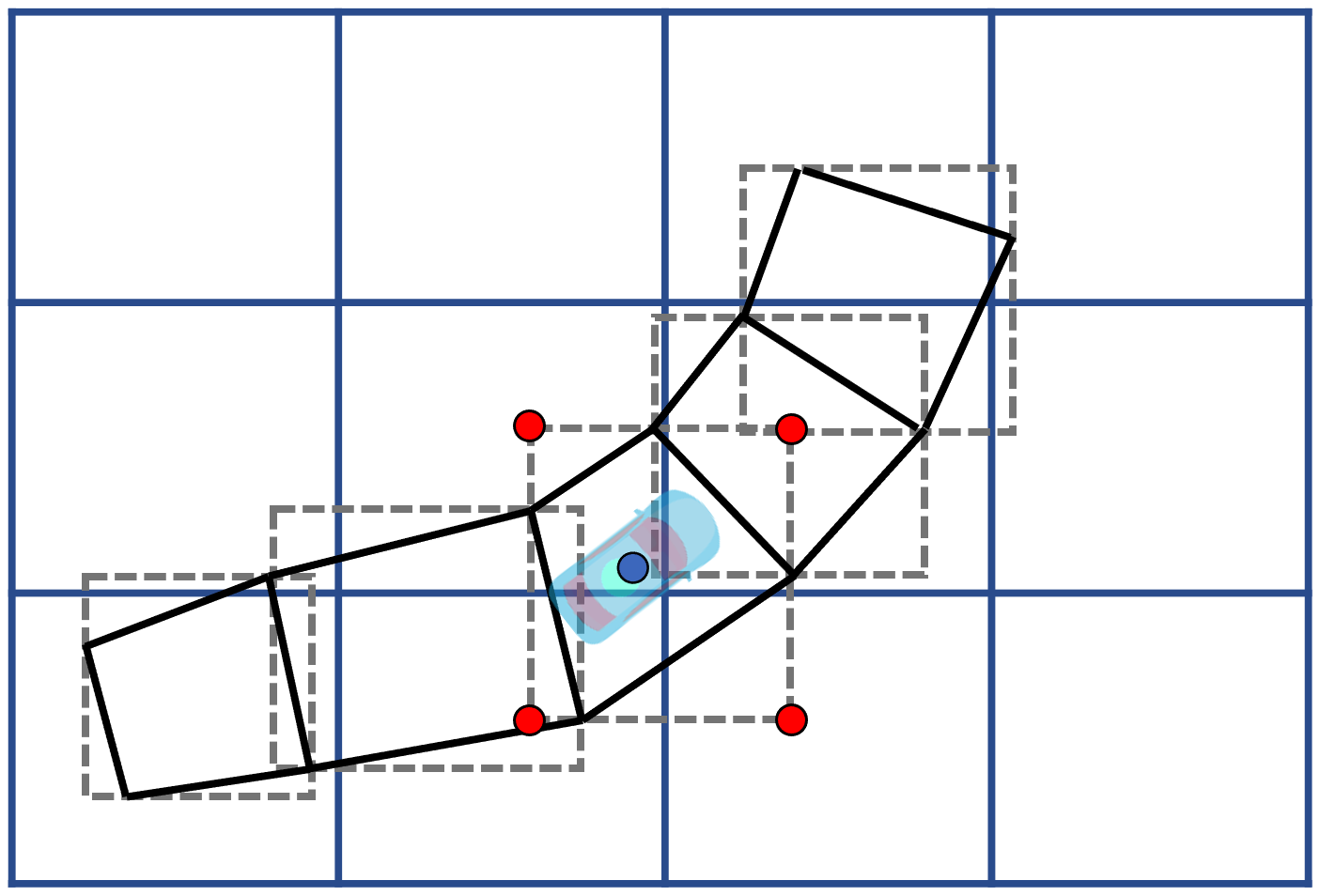}
\end{center}
\label{fig:images/quad_hash}
\caption{An exampling for understanding the hashing algorithm of the quadrilaterals of lanes}
\end{figure}
Lanes are static features with respect to a global reference point. Thus, if all lanes can be approximated before driving, or parts of lanes ahead of ego vehicle can be approximated before they come in the view range, the approximated parts can be used repeatedly during driving. In addition, if the quadrilaterals that approximate the lanes can be hashed, the time to search for the quadrilateral inside which the center or a corner position of a vehicle exists can be reduced from $\mathcal{O}(n)$ to $\mathcal{O}(1)$ where $n$ is the number of quadrilaterals to be searched for. Thus, we devise a hashing algorithm that can quickly search for quadrilaterals to which a given position can belong. Figure 6 is an example for understanding the hashing algorithm. Our hash function can hash a position in two-dimension to a bin, which can be represented as a blue rectangle of the smallest possible unit size. To store the information of quadrilaterals in bins, we hash the vertices of the rectangular bounding box that covers a quadrilateral and put the quadrilateral in the bin of each vertex so that the quadrilateral is stored in one to four bins. The grey boxes in the dashed line and the red dots are examples of the bounding boxes for quadrilaterals and the vertices of a bounding box, respectively. Practically, the duplicate storing of a quadrilateral in one bin is avoided for fast search. The width and height of any bin are kept longer than those of the bounding box that covers any quadrilateral, ensuring that vertices of any bounding box are hashed to the same bin or bins that are adjacent to each other. This also guarantees that for a position $p$, the hashing function $b = H(p)$ can give the bin that contains the quadrilateral to which $p$ belongs because $p$ is inside the bounding box that aligns with the bins. The blue dot in Figure 6 is the center position of the blue vehicle, inside the bounding box with the red dots as its vertices.
\\
\subsection{Inter-vehicle regions} \label{appendix:inter-vehicle_regions}
\begin{figure}[h]
\begin{center}
\includegraphics[width=1.0\textwidth]{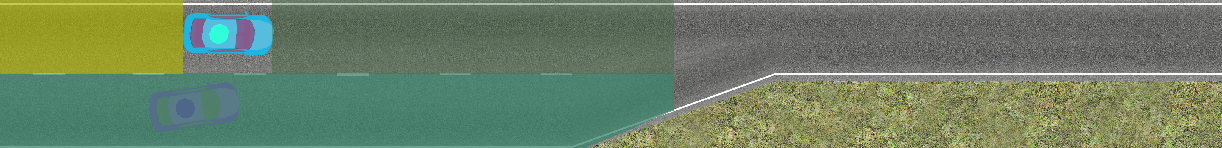}
\end{center}
\label{fig:IVRs}
\caption{An example of inter-vehicle regions for the purple-colored vehicle}
\end{figure}
An inter-vehicle region (IVR) in a lane is defined as the region between the rear end NQC, $n_{r} = (q_{r}, v_{r})$, and the front end NQC, $n_{f} = (q_{f}, v_{f})$, bordered by surrounding vehicles or any end of the view range. Figure 7 shows an example of IVRs for the purple-colored vehicle. Given normalized positions of both ends, the featuring global positions, ${(p_{0,l}, p_{0,r}), (p_{1,l}, p_{1,r}), \cdot, (p_{N-1,l}, p_{N-1,r}), (p_{N,l}, p_{N,r})}$, are acquired by sampling $N$ times with uniform distance interval on the left and right sides along the progressing direction, where $(p_{0,l}, p_{0,r})$ and $(p_{N,l}, p_{N,r})$ are defined by $n_{r}$ and $n_{f}$, respectively. By using the distance function \ref{appendix:NQC}, the distance between the rear and front $D_{total} = D(l, n_{r}, n_{f})$ is acquired, and from the rear, we can get the next sample at the distance of $D_{s} = D_{total}/(N-1)$ in turn.
\begin{table}[t]
\caption{Candidate behaviors with their available IVRs to pay mind to and the outline IVR}
\begin{center}
\begin{tabular}{lll}
    \toprule
    Behavioral mode                 & Set of available IVRs to pay mind to              & Outline IVR\\
    \midrule
    Stay in current IVR             & \{Current IVR\}, or \{Front IVR\} or \{Rear IVR\} & Current IVR\\ 
                                    & if a vehicle invades the lane of ego              &\\
    Maneuver to the other in lane   & \{Front IVR\} if ego is in rear IVR               & IVR in mind\\ 
    (optionally selectable)         & \{Rear IVR\} if ego is in front IVR               &\\
    Pay mind to the left            & \{IVRs on left lane to pay mind to                & Current IVR\\
                                    & if the lane exists, otherwise $\emptyset$         &\\
    Maneuver to the left            & \{IVRs on left lane ahead to pay mind to\}        & IVR in mind\\
                                    & if the lane exists, otherwise $\emptyset$         &\\
    Pay mind to the right           & \{IVRs on right lane to pay mind to\}             & Current IVR\\
                                    & if the lane exists, otherwise $\emptyset$         &\\
    Maneuver to the right           & \{IVRs on right lane ahead to pay mind to\}       & IVR in mind\\
                                    & if the lane exists, otherwise $\emptyset$         &\\
    \bottomrule
\end{tabular}
\end{center}
\label{tab:possible_combinations_for_candidate}
\end{table}
Table \ref{tab:possible_combinations_for_candidate} specifies candidate high-level commands of behaviors, the corresponding set of available IVRs to pay mind to, and the outline IVR.
\subsection{Other features, goal, and rewards} \label{appendix:other_state_features}

In our autonomous driving environment, on top of inter-vehicle regions, two additional types of state features exist: state of vehicles and range-sensing. 

The feature of a vehicle $j$ is 
\begin{equation}
s_{j} = [p_{fl,j}, p_{fr,j}, p_{rl,j}, p_{rr,j}, v_{j}, \psi_{j}]^{T},
\end{equation}
where $p_{fl,j} = (x_{fl,j}, y_{fl,j})$, $p_{fr,j} = (x_{fr,j}, y_{fr,j})$, $p_{rl,j} = (x_{rl,j}, y_{rl,j})$, and $p_{rr,j} = (x_{rr,j}, y_{rr,j})$.
are the positions of the front-left, front-right, rear-left, and rear-right corners, respectively,  $v_{j} = v_{x,j}, v_{y,j}$ is the velocity, and $psi_{j}$ is the orientation in radians. State of vehicles consists of features of ego vehicle and arbitrary number of surrounding vehicles on the road in the view range.

Range-sensing features are acquired by simulating range sensing of ego vehicle and processing the information about global or ego-centered positions. An odd number of rays are shot at uniform angular intervals, $\theta$, forming bilateral symmetry where the center ray is directed to the front of the vehicle. Figure 8 shows an example of rays shot from each vehicle. Total features of range-sensing include the positions of $N$ number of rays: the start, $r_{0}$ and the ends, ${r_{1}, r_{2}, \cdot, r_{N-1}, r_{N}}$. Rays end where they hit obstacles such as other vehicles or lane shoulders, or at the maximum distance, $d_{max}$. We used rays with $N = 25$, $N \theta = 120$, and $d_{max}=1000$.
\begin{figure}[h]
\begin{center}
\includegraphics[width=0.5\textwidth]{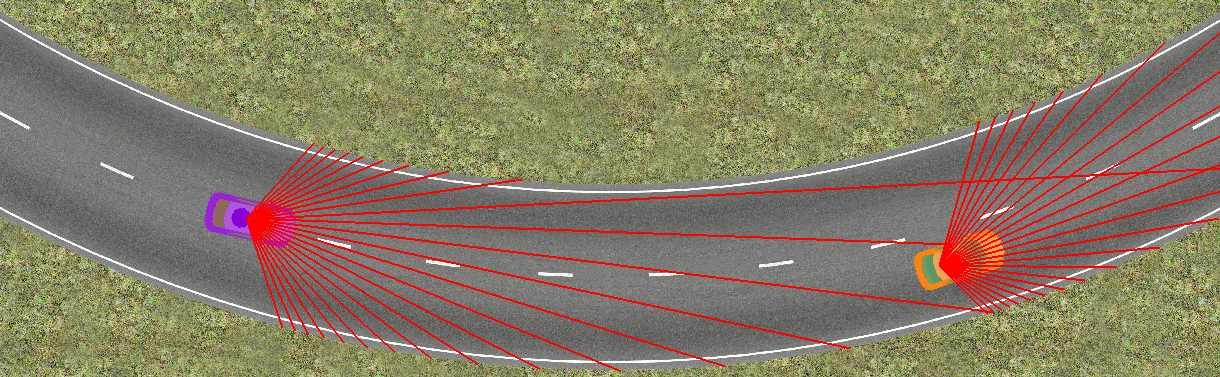}
\end{center}
\label{fig:range_sensing}
\caption{Range sensing of vehicles}
\end{figure}
\begin{table}[t]
\caption{Reward types and their definition given in our environment}
\label{tab:reward_types}
\begin{center}
\begin{tabular}{ll}
    \toprule
    Reward type        & Definition\\
    \midrule
    Goal reward        & $1$ if the ego center position gets into the long-term goal\\ 
    Collision penalty  & $-1$ if the vehicle image overlays on\\
                       & the lane shoulders or other vehicles\\
    Progression reward & $c_{pro} v_{quad}$ where $0 < c_{pro} < 1$\\
    Max speed penalty  & $c_{max} min(v_{quad} - l_{max}, 0)$\\
                       & for speed limit $l_{max}$ where $-1 < c_{max} < -c_{pro}$\\
    Min speed penalty  & $c_{min} max(l_{min} - v_{quad}, 0)$\\
                       & for speed limit $l_{min}$ where $-1 < c_{min} < 0$\\
    \bottomrule
\end{tabular}
\end{center}
\end{table}
Inter-vehicle regions can be applicable as the long-term goal.
For experiments, we simply define a long-term goal region as the rectangular box $(l, r, t, b)$ to deal with two types of goal regions, the region reached after the right end of a random lane and all lanes. Including the goal reaching reward, four types of rewards and their definitions are as in Table \ref{tab:reward_types}, where $v_{quad}$ is the velocity along the progressing direction of the quadrilateral that the agent belongs to, $l_{max}$ and $v_{min}$ are maximum and minimum speed limits, respectively, and $c_{pro}$, $c_{max}$, and $c_{min}$ are constants.

\end{document}